\documentclass[10pt,leqno]{amsart}
\usepackage{graphicx}
\usepackage{indentfirst,csquotes}

\usepackage{amssymb,amsthm,amsmath}
\usepackage{xcolor,paralist,hyperref,fancyhdr,etoolbox}

\hypersetup{ colorlinks=true, linkcolor=black, filecolor=black, urlcolor=black }

\usepackage{lipsum}

\begin{document}
\title{An Accurate and Interpretable Hyper Graph Neural Network for GBM Survival Prediction} 
\author[Mushahid Intesum]{Mushahid Intesum}
\date{\today}
\address{Independent Researcher}
\email{intesummushahid@gmail.com}
\maketitle

\let\thefootnote\relax
\footnotetext{} 

\begin{abstract}
Survival prediction for glioblastoma multiforme (GBM) demands models that are both accurate and interpretable, yet existing approaches treat these objectives as competing, where performant models sacrifice transparency, while interpretable models accept degraded predictive power. We argue that this trade-off is not inherent. Graph neural networks offer a structural foundation for extracting interpretable, explainable representations without compromising discriminative ability. Furthermore, current methods typically rely on a single imaging modality, underutilizing the complementary information available across multi-modal MRI and clinical metadata. We propose a multi-modal framework that integrates three components to address both objectives simultaneously: (1)~a \textit{sheaf hypergraph neural network} that captures higher-order relationships among tissue patches through directional, asymmetric message passing; (2)~a \textit{concept bottleneck layer} that compresses learned representations into clinically grounded concepts, enforcing ante-hoc interpretability; and (3)~an \textit{extension sufficiency test (EST) regularizer} that penalizes unfaithful explanations during training, ensuring that model explanations genuinely reflect the internal decision process. Clinical and genomic features are incorporated through gated fusion, preserving the dominant prognostic signal of molecular markers while retaining concept-level traceability. Evaluated on 593 patients from the UPenn-GBM dataset under 5-fold cross-validation, our framework achieves a concordance index of 0.643 with the lowest fold-level variance among all compared models (std = 0.015). To our knowledge, this is the first work to unify sheaf hypergraph convolution, concept bottleneck supervision, and EST regularization for interpretable survival prediction from brain MRI.
\end{abstract}

\section{Introduction}

Glioblastoma multiforme (GBM) is the most aggressive primary brain tumor, with a median survival of approximately 14 months under the standard Stupp protocol of maximal surgical resection followed by concurrent temozolomide and radiotherapy \cite{stupp2005}. Treatment decisions in GBM are inherently time-sensitive and consequential: the choice between aggressive re-intervention, experimental therapy enrollment, and palliative care depends critically on individualized prognosis. Accurate survival prediction can inform these decisions, yet clinicians are unlikely to act on predictions they cannot understand, creating a fundamental tension between model performance and clinical trust.

Current deep learning models increasingly leverage multi-modal patient data, combining imaging, genomic markers, and clinical metadata, to achieve strong predictive performance. However, these predictions are often opaque, offering no insight into which imaging biomarkers, tumor characteristics, or molecular features drive the risk assessment. In clinical oncology, where treatment decisions carry significant consequences, understanding \textit{why} a patient is classified as high-risk is as important as the prediction itself. Regulatory trends, including the EU AI Act and FDA guidelines on clinical decision support, further underscore the need for models that can articulate their reasoning in terms that align with clinical knowledge.

GBM presents a particularly compelling case for structured, spatial modeling. The tumor microenvironment is spatially heterogeneous: contrast-enhancing regions indicate blood-brain barrier disruption, FLAIR hyperintensities delineate peritumoral edema, and DTI abnormalities reveal white matter infiltration. These tissue properties carry distinct prognostic significance and interact in complex, multi-way patterns. For instance, the spatial relationship between an enhancing rim, a necrotic core, and surrounding edema collectively characterizes tumor aggressiveness in ways that individual features in isolation cannot capture. Fixed feature vector approaches such as DeepSurv \cite{DeepSurv} and Cox-nnet \cite{cox-net} collapse this spatial structure into a single vector, discarding the relational information that makes imaging clinically valuable.

Graph neural networks offer a natural framework for encoding spatial tissue topology, representing image patches as nodes and their spatial relationships as edges. However, standard pairwise graphs only capture local adjacency and cannot model the higher-order, multi-way interactions among tissue subregions. Hypergraph formulations address this limitation by allowing a single hyperedge to connect groups of related nodes, representing multi-way tissue relationships such as co-occurring enhancement patterns across MRI modalities. Methods such as MRePath \cite{mre-path} demonstrate the effectiveness of hypergraph construction for cancer survival prediction, but their learned representations remain opaque, with no mechanism for tracing a prediction back to specific clinical concepts.

Concept Bottleneck Models (CBMs) \cite{cbm-koh} provide a complementary approach by routing predictions through a layer of human-interpretable concept activations. This design provides ante-hoc interpretability: the model must express its reasoning through clinically meaningful intermediate variables. However, this interpretability comes at a cost. Compressing rich embeddings through a small set of concepts creates an information bottleneck that typically reduces prediction accuracy. Whether this trade-off is acceptable depends on whether downstream components can recover the lost performance. Additionally, existing medical CBMs such as HyperCBM \cite{hypercbm} target classification tasks and have not been adapted for survival analysis or integrated with multi-modal clinical fusion.

A further challenge lies in the faithfulness of explanations. Post-hoc explanation methods for GNNs, such as GNNExplainer \cite{gnn-explainer}, can produce explanations that appear clinically plausible but do not actually reflect the model's internal reasoning \cite{se-gnn-audit}. A clinician acting on an unfaithful explanation is arguably worse off than one with no explanation at all. This motivates training-time mechanisms that ensure explanations are not merely plausible but verified to represent the model's actual decision process.

In this work, we propose a unified framework that addresses these challenges jointly. We combine sheaf hypergraph learning, concept bottleneck interpretability, multi-modal clinical fusion, and training-time faithfulness regularization for GBM survival prediction. Our contributions are as follows:

\begin{enumerate}
    \item We build a sheaf hypergraph concept bottleneck GNN that integrates multi-modal MRI (T1, T2, FLAIR, DTI, Perfusion) with clinical and genomic metadata (IDH1, MGMT, KPS) for GBM survival prediction, with all predictions traceable through 8 clinically grounded imaging concepts.
    \item We introduce EST-regularized training that improves both explanation faithfulness and cross-fold stability, reducing fold variance from 0.027 to 0.015 in standard deviation.
    \item We conduct a 7-configuration progressive ablation study on 593 UPenn-GBM patients, demonstrating that the concept bottleneck imposes only a $-3.7\%$ interpretability cost while clinical fusion provides $+18.2\%$, the single largest performance gain.
    \item We achieve a concordance index of $0.643 \pm 0.015$ with fully traceable predictions, where each survival estimate can be decomposed into specific concept activations and modality contributions.
\end{enumerate}

To our knowledge, this is the first framework to combine sheaf hypergraph learning, concept bottleneck interpretability, and multi-modal clinical fusion for brain tumor survival prediction with faithful, auditable explanations.

\section{Related Work}

\subsection{Neural Survival Models}

The Cox proportional hazards model \cite{cox-ph} models the hazard function as a log-linear combination of covariates. DeepSurv \cite{DeepSurv} and Cox-nnet \cite{cox-net} replace the linear predictor with deep networks, enabling nonlinear interactions while preserving the partial likelihood objective. DeepHit \cite{deephit} and Nnet-survival \cite{nnet-survival} adopt discrete-time formulations that directly estimate survival probabilities at predefined intervals, eliminating the proportional hazards assumption. Our framework uses the discrete-time negative log-likelihood, providing calibrated survival estimates at each time bin. However, all of these methods operate on fixed feature vectors and cannot model spatial or relational structure within imaging data.

\subsection{Graph and Hypergraph Neural Networks}

Graph neural networks model tissue topology by treating image patches or cells as nodes with spatial edges \cite{patch-gcn, cgc-net, hact-net}. Patch-GCN \cite{patch-gcn} applies graph convolutions over kNN patch graphs for survival prediction, preserving spatial context lost in attention-based MIL methods \cite{attention-mil, clam}. Hypergraphs generalize this by allowing edges to connect arbitrary node subsets. HGNN \cite{hgnn} introduced spectral hypergraph convolution, HyperGCN \cite{hyper-gcn} approximates it through Laplacian expansions, and AllSet \cite{allset} provides a unified set-function framework. Sheaf neural networks \cite{sheaf-nn, sheaf-hgnn} further enrich hypergraph structure with learned restriction maps on each node-hyperedge incidence, enabling heterogeneous message passing. Our SheafHGNN uses this formulation with a dual-space (spatial + feature) hypergraph construction. MRePath \cite{mre-path} is most related, constructing hypergraphs over image patches with cross-modality rebalancing for survival prediction; our framework extends this with concept supervision and faithfulness regularization.

\subsection{Concept Bottleneck Models}

CBMs \cite{cbm-koh} route predictions through interpretable concept activations, providing ante-hoc explanations at the cost of an information bottleneck. Concept Embedding Models \cite{cem} mitigate this cost by learning richer concept representations. Label-free \cite{label-free-cbm} and post-hoc \cite{posthoc-cbm} variants reduce annotation requirements, while interactive CBMs \cite{interactive-cbm} enable test-time concept correction. HyperCBM \cite{hypercbm} extends CBMs to hypergraph-structured medical data. However, existing medical CBMs target classification rather than survival analysis. Our framework integrates a CBM into a survival pipeline with HECRL for inter-concept refinement.

\subsection{Multimodal Fusion and Explainability}

Attention-based fusion methods such as MCAT \cite{mcat} and Porpoise \cite{porpoise} learn cross-modal interactions between pathology and genomics for survival prediction. Our framework uses gated attention fusion \cite{gated-fusion} between concept-pooled imaging features and clinical covariates.

Post-hoc GNN explanation methods identify task-relevant substructures: GNNExplainer \cite{gnn-explainer} learns soft masks over edges, PGExplainer \cite{pgexplainer} trains parameterized explainers, and SubgraphX \cite{subgraphx} searches for important subgraphs via Shapley values. However, Azzolin et al. \cite{se-gnn-audit} demonstrate that many GNN explanations fail rigorous faithfulness tests.

\section{Background}

\subsection{Graph Neural Networks}

Graph Neural Networks (GNNs) operate on graph-structured data through message passing. Given a graph $G = (V, E)$ with node features $\mathbf{x}_v$, a GNN layer updates each node's representation by aggregating neighbor information:

\begin{equation}
    \mathbf{h}_v^{(l+1)} = \text{UPDATE}\left(\mathbf{h}_v^{(l)}, \text{AGG}\left(\{\mathbf{h}_u^{(l)} : u \in \mathcal{N}(v)\}\right)\right)
\end{equation}

After $L$ layers, node representations encode information from their $L$-hop neighborhood, capturing spatial relationships between tissue regions relevant to tumor prognosis.

\subsection{Hypergraphs and Sheaf Hypergraphs}

A hypergraph $\mathcal{H} = (V, \mathcal{E})$ generalizes standard graphs by allowing each hyperedge $e \in \mathcal{E}$ to connect arbitrarily many nodes, naturally modeling groups of tissue patches sharing biological properties. Message passing follows a two-step node-to-hyperedge-to-node process \cite{hyper-gcn}:

\begin{equation}
    \mathbf{h}_v^{(l+1)} = \sigma\left(\sum_{e \in \mathcal{E}(v)} \frac{1}{|e|} \sum_{u \in e} \mathbf{W}^{(l)} \mathbf{h}_u^{(l)}\right)
\end{equation}

Sheaf hypergraphs \cite{sheaf-hgnn} extend this with learned linear maps $\mathbf{F}_{v \trianglelefteq e} : \mathbb{R}^d \rightarrow \mathbb{R}^d$ for each node-hyperedge pair, enabling directional and asymmetric information flow. This allows different nodes to contribute different aspects of their representation to the same hyperedge---for example, a tumor core patch and a peritumoral patch can contribute differently to a shared tissue neighborhood.

\subsection{Concept Bottleneck Models}

Concept Bottleneck Models (CBMs) \cite{cbm-koh} route all information through $K$ human-interpretable concepts before prediction. Given input $\mathbf{x}$, a CBM predicts concept values $\mathbf{c} = f(\mathbf{x}) \in \mathbb{R}^K$, then computes $\hat{y} = g(\mathbf{c})$ solely from these concepts. Since $g$ has no access to raw features, every prediction is inherently traceable through concept activations---providing \textit{ante-hoc} interpretability, as opposed to \textit{post-hoc} methods like GNNExplainer \cite{gnn-explainer}.

\subsection{Survival Analysis}

Survival analysis predicts time-to-event while handling \textit{right censoring}: patients alive at study end have survival times only known to exceed the observation period. We adopt a discrete-time formulation that predicts per-bin hazard probabilities $\hat{h}_k = P(\text{event in bin } k \mid \text{survived to bin } k)$, trained with negative log-likelihood loss. Performance is measured by Harrell's concordance index (C-Index) \cite{harrell-c-index}, where 0.5 is random, 1.0 is perfect, and 0.60--0.70 is typical for GBM.

\subsection{Explainability and Faithfulness}

A \textit{faithful} explanation identifies the subgraph that accurately represents the model's internal decision process: used as the sole input, it should reproduce the original prediction \cite{se-gnn-audit}. We evaluate faithfulness using \textbf{Sufficiency} (whether the explanation subgraph alone reproduces the prediction), \textbf{Fidelity-minus} (prediction shift when retaining only the explanation), and the \textbf{Extension Sufficiency Test (EST)}, which detects anchor-set degeneracy by testing whether non-explanation nodes affect the prediction. We use EST both for evaluation and as a training-time regularizer.

\section{Methodology}

\begin{figure*}[t]
\centering
\includegraphics[width=\textwidth]{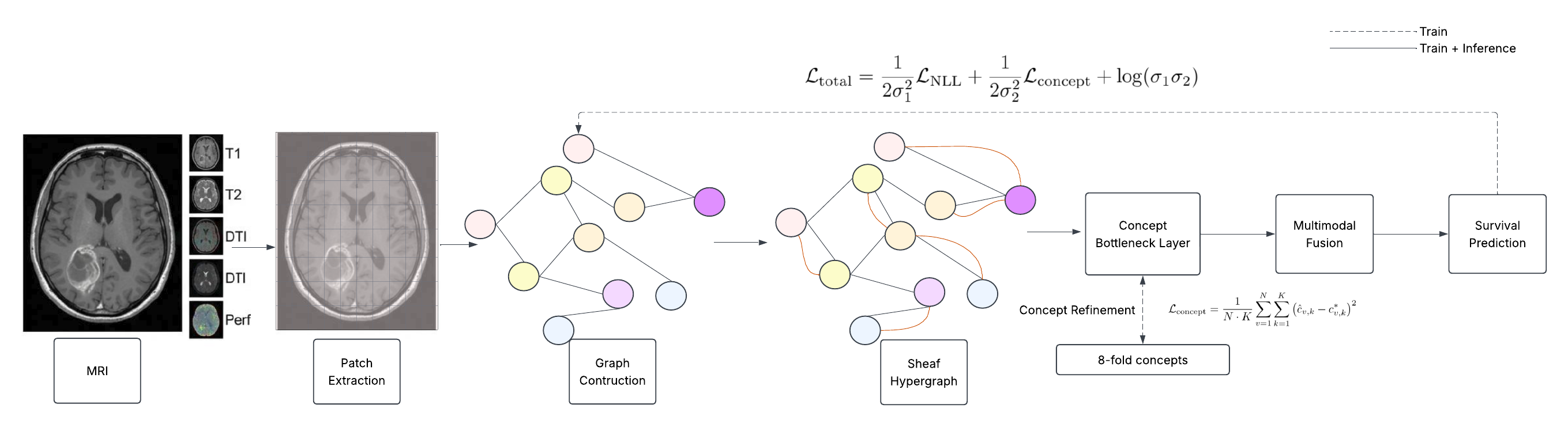}
\caption{Architecture overview. Multi-modal MRI patches are encoded into a dual-space sheaf hypergraph, processed by SheafHGNN, compressed through a concept bottleneck (8 concepts), refined by HECRL, and fused with clinical features via gated attention for survival prediction. EST regularization (training only) penalizes prediction shifts when non-explanation nodes are added.}
\label{fig:architecture}
\end{figure*}

We present a unified framework for interpretable GBM survival prediction that integrates multi-modal brain MRI, clinical metadata, and genomic markers through a sheaf hypergraph concept bottleneck architecture. The pipeline consists of five stages: (1) data preprocessing and hypergraph construction, (2) sheaf hypergraph message passing, (3) concept bottleneck prediction, (4) multi-modal fusion, and (5) survival prediction. Figure~\ref{fig:architecture} provides an overview of the full pipeline.

\subsection{Data Preprocessing}

\subsubsection{MRI Patch Extraction}

Given a patient's multi-modal brain MRI consisting of six modalities (T1-pre, T1-post, T2, FLAIR, DTI, Perfusion), we extract $N$ non-overlapping patches of size $16 \times 16$ from each axial slice. Each patch is flattened across all modalities to produce a feature vector $\mathbf{x}_i \in \mathbb{R}^{1536}$ (i.e., $6 \times 16 \times 16 = 1536$). Background patches (those with mean intensity below a threshold) are discarded. The remaining patches form the node set $V = \{v_1, \ldots, v_N\}$ of the patient graph, where each node corresponds to a localized tissue region.

\subsubsection{Concept Ground Truth Computation}

For each patch, we precompute 8 imaging concept values directly from raw intensities without requiring segmentation ground truth. These serve as supervision targets for the concept bottleneck:

\begin{enumerate}
    \item \textbf{Enhancement ratio} ($c_1$): $\log(1 + \text{clip}(\text{T1-post} / (\text{T1-pre} + \epsilon), 0, 10))$, measuring blood-brain barrier breakdown in contrast-enhancing tumor regions.
    \item \textbf{FLAIR z-score} ($c_2$): $\mu_{\text{FLAIR}} / \sigma_{\text{FLAIR}}$ within the patch, indicating peritumoral edema.
    \item \textbf{T2 abnormality} ($c_3$): T2 and FLAIR interaction term, capturing non-enhancing tumor extent.
    \item \textbf{DTI mean diffusivity} ($c_4$): Mean DTI signal within the patch, reflecting white matter disruption.
    \item \textbf{DTI FA proxy} ($c_5$): Coefficient of variation of DTI signal, approximating fractional anisotropy as a measure of axonal integrity.
    \item \textbf{Intensity heterogeneity} ($c_6$): Standard deviation across all modalities within the patch, capturing intratumoral heterogeneity.
    \item \textbf{Boundary complexity} ($c_7$): Left unsupervised (graph-learned), intended to capture tumor margin irregularity through message passing.
    \item \textbf{Spatial location} ($c_8$): Normalized z-coordinate, encoding the patch's depth within the brain volume.
\end{enumerate}

All concept values are standardized to zero mean and unit variance across the dataset.

\subsubsection{Clinical Features}

We extract 18 clinical features from the UPenn-GBM metadata: age at diagnosis, gender, IDH1 mutation status, MGMT methylation status, Karnofsky Performance Score (KPS), gross total resection status (GTR), and additional molecular and demographic attributes. Missing values are imputed with median values and accompanied by binary missingness indicator flags.

\subsection{Dual-Space Hypergraph Construction}

We construct a patient-specific hypergraph $\mathcal{H} = (V, \mathcal{E}_T \cup \mathcal{E}_F)$ using two complementary strategies following MRePath \cite{mre-path}:

\subsubsection{\textbf{Topological hyperedges}} ($\mathcal{E}_T$): For each patch $v_i$, we form a hyperedge containing all patches within a spatial radius of $\delta = 2.5$ patch-grid units in normalized coordinate space. This captures local tissue neighborhoods where adjacent patches share a spatial context. Hyperedges with fewer than 2 members are discarded and those exceeding 12 members are pruned to the nearest neighbors.

\subsubsection{\textbf{Feature hyperedges}} ($\mathcal{E}_F$): For each patch $v_i$, we form a hyperedge containing the $k = 9$ patches with highest cosine similarity in concept feature space. This captures non-local structural similarity, allowing distant patches with similar imaging characteristics (e.g., two separate enhancing regions) to share information.

The union $\mathcal{E} = \mathcal{E}_T \cup \mathcal{E}_F$ forms the complete hyperedge set. Type labels are preserved to allow the model to distinguish between spatial and feature-based groupings. The hypergraph is stored as an incidence index $\mathbf{I} \in \mathbb{R}^{2 \times |\mathcal{E}|}$ mapping node indices to hyperedge indices.

\subsection{Sheaf Hypergraph Neural Network}

\subsubsection{Patch Encoder}

Raw patch features $\mathbf{x}_i \in \mathbb{R}^{1536}$ are projected into a $d$-dimensional embedding space ($d = 64$) using a three-layer MLP with LayerNorm and GELU activations:

\begin{equation}
    \mathbf{h}_i^{(0)} = \text{PatchEncoder}(\mathbf{x}_i) \in \mathbb{R}^{d}
\end{equation}

\subsubsection{Sheaf Message Passing}

We apply $L = 3$ layers of sheaf hypergraph message passing. Each layer $l$ performs the following:

\textbf{Step 1: Vertex to Hyperedge.} Node features are transformed through a learned sheaf map $\mathbf{F}_{v \trianglelefteq e}$ and aggregated within each hyperedge by averaging:

\begin{equation}
    \mathbf{m}_e^{(l)} = \frac{1}{|e|} \sum_{v \in e} \mathbf{F}_{v \trianglelefteq e}^{(l)} \mathbf{h}_v^{(l)}
\end{equation}

where $\mathbf{F}_{v \trianglelefteq e}^{(l)} \in \mathbb{R}^{d \times d}$ is a learned linear map shared across all node-hyperedge pairs within each layer. This sheaf map controls how each node's features are projected when contributing to a hyperedge, enabling direction-aware information flow.

\textbf{Step 2: Hyperedge to Vertex.} Hyperedge features are transformed through a second sheaf map $\mathbf{F}_{e \trianglelefteq v}$ and propagated back to member nodes:

\begin{equation}
    \tilde{\mathbf{h}}_v^{(l)} = \sum_{e \in \mathcal{E}(v)} \mathbf{F}_{e \trianglelefteq v}^{(l)} \mathbf{m}_e^{(l)}
\end{equation}

\textbf{Step 3: Sheaf Laplacian Normalization.} The aggregated messages are normalized by the inverse square root of node degree and passed through a weight matrix, LayerNorm, and GELU activation:

\begin{equation}
    \mathbf{h}_v^{(l+1)} = \text{GELU}\left(\text{LayerNorm}\left(\mathbf{W}^{(l)} \cdot \mathbf{D}_v^{-1/2} \tilde{\mathbf{h}}_v^{(l)}\right)\right)
\end{equation}

where $\mathbf{D}_v$ is the node degree (number of incident hyperedges). Residual connections are applied between layers when input and output dimensions match.

\subsection{Concept Bottleneck}

\subsubsection{Concept Prediction}

After $L$ layers of sheaf message passing, node embeddings $\mathbf{h}_v \in \mathbb{R}^d$ are passed through a concept predictor that produces 8 concept values per node. The predictor consists of a shared trunk followed by 8 disentangled prediction heads, one per concept:

\begin{equation}
    \hat{c}_{v,k} = \phi_k\left(\text{SharedTrunk}(\mathbf{h}_v)\right), \quad k = 1, \ldots, 8
\end{equation}

where each $\phi_k$ is a two-layer MLP. The concept predictions are supervised with MSE loss against the precomputed ground truth values:

\begin{equation}
    \mathcal{L}_{\text{concept}} = \frac{1}{N \cdot K} \sum_{v=1}^{N} \sum_{k=1}^{K} \left(\hat{c}_{v,k} - c_{v,k}^*\right)^2
\end{equation}

where $K = 7$ (concept $c_7$ is excluded from supervision as it is graph-learned).

\subsubsection{HECRL: Inter-Concept Refinement}

Following HyperCBM \cite{hypercbm}, we apply a Hyperedge Concept Refinement Layer (HECRL) that uses multi-head self-attention across the 8 concept dimensions. This captures inter-concept dependencies such as the correlation between high enhancement ratio ($c_1$) and high FLAIR signal ($c_2$), which together indicate the edema pattern characteristic of aggressive tumors:

\begin{equation}
    \hat{\mathbf{c}}_v' = \text{MHSA}(\hat{\mathbf{c}}_v) + \hat{\mathbf{c}}_v
\end{equation}

The refined concept vectors $\hat{\mathbf{c}}_v' \in \mathbb{R}^K$ form the bottleneck output. All downstream modules see only these concept values, creating a hard information barrier that ensures prediction traceability.

\subsubsection{Graph Pooling}

Node-level concept representations are pooled to a single graph-level embedding using attention-weighted aggregation:

\begin{equation}
    \mathbf{g} = \sum_{v=1}^{N} \alpha_v \cdot \text{MLP}(\hat{\mathbf{c}}_v')
\end{equation}

where $\alpha_v = \text{softmax}(\mathbf{w}^\top \hat{\mathbf{c}}_v')$ are learned attention weights that identify the most diagnostically relevant patches.

\subsection{Multi-Modal Fusion}

We fuse the graph-level imaging embedding $\mathbf{g} \in \mathbb{R}^d$ with clinical features using a dynamic weighting scheme inspired by MRePath \cite{mre-path}.

\subsubsection{Clinical Encoder}

Clinical features $\mathbf{x}_{\text{clin}} \in \mathbb{R}^{18}$ are projected into the same latent space as imaging through a two-layer MLP:

\begin{equation}
    \mathbf{g}_{\text{clin}} = \text{ClinicalEncoder}(\mathbf{x}_{\text{clin}}) \in \mathbb{R}^d
\end{equation}

\subsubsection{Dynamic Modality Rebalancing}

Per-patient weights are computed using mono-confidence scores that estimate each modality's standalone reliability:

\begin{equation}
    w_{\text{img}}, w_{\text{clin}} = \text{softmax}\left(\Phi_{\text{img}}(\mathbf{g}),\, \Phi_{\text{clin}}(\mathbf{g}_{\text{clin}})\right)
\end{equation}

where $\Phi_{\text{img}}$ and $\Phi_{\text{clin}}$ are learned linear projections. This prevents modality collapse: if clinical data is heavily missing or unreliable for a particular patient, the model automatically upweights imaging, and vice versa.

The fused representation is computed as a weighted combination followed by cross-attention alignment:

\begin{equation}
    \mathbf{f} = w_{\text{img}} \cdot \mathbf{g} + w_{\text{clin}} \cdot \mathbf{g}_{\text{clin}} + \text{CrossAttn}(\mathbf{g}, \mathbf{g}_{\text{clin}})
\end{equation}

\subsection{Survival Prediction}

\subsubsection{Survival Head}

The fused embedding $\mathbf{f} \in \mathbb{R}^d$ is passed through a three-layer MLP with dropout to produce hazard logits for $K = 4$ discrete time bins:

\begin{equation}
    \hat{\mathbf{h}} = \text{SurvivalHead}(\mathbf{f}) \in \mathbb{R}^{K}
\end{equation}

Time bin boundaries are computed from training set event times using quartiles. Hazard probabilities are obtained via $\hat{h}_k = \sigma(\hat{\mathbf{h}}_k)$.

\subsubsection{Loss Function}

The total training loss combines survival prediction, concept supervision, and learned task weighting using the Kendall uncertainty framework \cite{kendall-uncertainty}:

\begin{equation}
    \mathcal{L}_{\text{total}} = \frac{1}{2\sigma_1^2} \mathcal{L}_{\text{NLL}} + \frac{1}{2\sigma_2^2} \mathcal{L}_{\text{concept}} + \log(\sigma_1 \sigma_2)
\end{equation}

where $\log \sigma_1^2$ and $\log \sigma_2^2$ are learned parameters, and $\mathcal{L}_{\text{NLL}}$ is the discrete-time negative log-likelihood survival loss described in Section 2.4.

\subsubsection{Pairwise Concordance Ranking Loss}

In addition to the NLL loss, we add a pairwise ranking loss that directly optimizes the concordance ordering. For each gradient accumulation step, the current patient's predicted risk is compared against a buffer of recently seen patients:

\begin{equation}
    \mathcal{L}_{\text{rank}} = -\frac{1}{|\mathcal{P}|} \sum_{(i,j) \in \mathcal{P}} \log \sigma\left(r_i - r_j\right)
\end{equation}

where $\mathcal{P}$ is the set of concordant pairs (patient $i$ died before patient $j$) and $r_i = \sum_k \hat{h}_k^{(i)}$ is the cumulative risk score. The ranking loss is added with weight $\lambda_{\text{rank}} = 0.5$.

\subsection{EST Regularization}

Following the SE-GNN auditing framework \cite{se-gnn-audit}, we use the Extension Sufficiency Test not only as an evaluation metric but as a training-time regularizer. After an initial warmup of 3 epochs, we periodically compute the EST loss for a subset of training samples:

\begin{equation}
    \mathcal{L}_{\text{EST}} = \left\| \hat{\mathbf{h}}(G_R) - \hat{\mathbf{h}}(G_{R \cup S}) \right\|_2
\end{equation}

where $G_R$ is the explanation subgraph (top 20\% of nodes by concept activation magnitude) and $G_{R \cup S}$ is the explanation extended with a randomly sampled subset $S$ of non-explanation nodes. This penalizes the model when adding non-explanation nodes changes the prediction, encouraging the model to concentrate its reasoning within the explanation subgraph. The EST loss is added to the total loss with weight $\lambda_{\text{EST}} = 0.1$.

\section{Experimental Results}

\subsection{Experimental Setup}

\subsubsection{Dataset}

We evaluate on the UPenn-GBM dataset \cite{upenn-gbm}, which contains multi-modal brain MRI scans with paired clinical metadata and survival outcomes. After filtering for patients with complete modality coverage (T1-pre, T1-post, T2, FLAIR, and at least one of DTI or Perfusion), 593 of 630 patients remain. Survival times range from 1 to 4,752 days, with 62\% of patients experiencing the event (deceased) and 38\% right-censored.

\subsubsection{Training Configuration}

We train with AdamW \cite{adamw} (learning rate $10^{-4}$, weight decay $10^{-5}$) using a sequential learning rate schedule: linear warmup from $10^{-6}$ to $10^{-4}$ over 3 epochs, followed by cosine annealing. Gradient accumulation with step size 4 is used to simulate larger batch sizes given the single-patient-per-step constraint of variable-size graphs. Gradients are clipped to a maximum norm of 1.0.

Early stopping with patience of 7 epochs monitors validation C-Index, halting training when the metric fails to improve. Maximum training length is set to 30 epochs. All experiments use stratified 5-fold cross-validation. The embedding dimension is $d =64$, and the total model contains approximately 1.07M parameters.

\subsection{Model Configurations}

To isolate the contribution of each architectural component, we define seven experiment configurations that progressively add modules:

\begin{table}[h]
\centering
\begin{tabular}{c l l}
\hline
\textbf{Exp} & \textbf{Components} \\
\hline
E1 & kNN graph + GNN + SurvivalHead \\
E2  & Dual-space hypergraph + SheafHGNN \\
E3  & E2 + CBM + HECRL \\
E4  & E3 + Multi-modal fusion \\
E5  & E4 + Multi-granular tree pooling \\
E6  & E4 + EST training-time regularization \\
E7  & E4 + TIF Tree + EST Regularizer  \\
\hline
\end{tabular}
\caption{Progressive ablation configurations. E1 through E4 build the core pipeline, while E5, E6, and E7 evaluate auxiliary modules.}
\label{tab:ablation-configs}
\end{table}

E1 uses a simple kNN graph with standard GNN message passing (no hypergraph, no concepts). E2 replaces the kNN graph with the dual-space sheaf hypergraph. E3 adds the concept bottleneck and HECRL. E4 adds clinical fusion. E5 and E6 independently add the TIF tree and EST regularizer on top of E4, while E7 combines both.

\subsection{Ablation Study}

Table~\ref{tab:ablation-results} presents the full ablation study across all seven configurations.

\begin{table}[h]
\centering
\begin{tabular}{c l c c c}
\hline
\textbf{Exp}  & \textbf{C-Index} & \textbf{Std} & \textbf{Params} \\
\hline
E1  & 0.5519 & 0.012 & 1,026,992 \\
E2  & 0.5601 & 0.016 & 1,026,992 \\
E3 & 0.5391 & 0.019 & 1,027,281 \\
E4  & 0.6373 & 0.027 & 1,074,963 \\
E5  & 0.6392 & 0.041 & 1,167,131 \\
E6& 0.6431 & 0.015 & 1,074,963 \\
E7 & 0.6435 & 0.023 & 1,167,131 \\
\hline
\end{tabular}
\caption{5-fold cross-validation results on 593 UPenn-GBM patients. }
\label{tab:ablation-results}
\end{table}

\begin{figure}[t]
\centering
\includegraphics[width=\columnwidth]{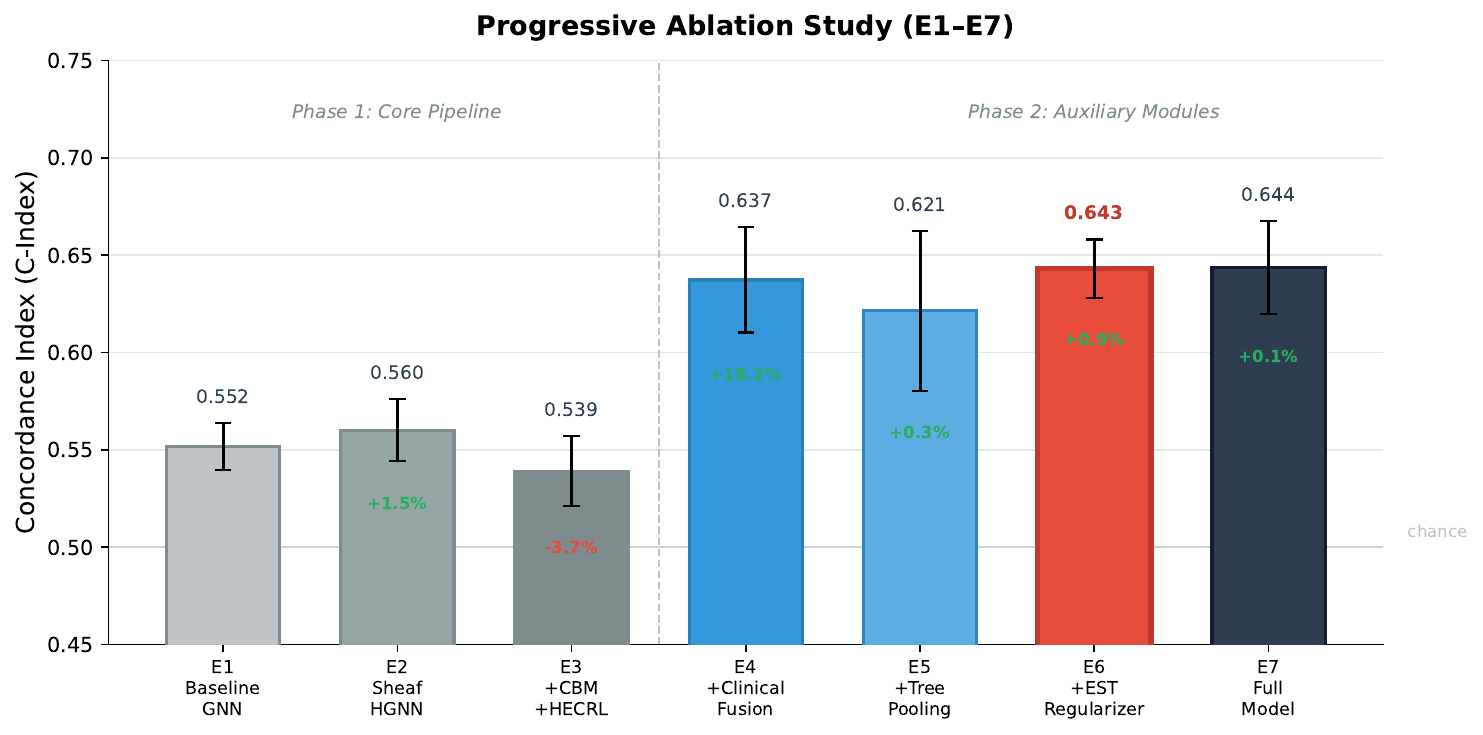}
\caption{Progressive ablation study (E1--E7). Phase~1 builds the core pipeline; Phase~2 evaluates auxiliary modules atop E4. Error bars denote standard deviation across 5 folds. Clinical fusion (E3$\rightarrow$E4) contributes the largest single gain (+18.2\%). E6 (red) achieves the best accuracy--stability trade-off.}
\label{fig:ablation}
\end{figure}

\subsubsection{Component Contributions}

Table~\ref{tab:ablation-deltas} quantifies the marginal contribution of each component.

\begin{table}[h]
\centering
\begin{tabular}{l c l}
\hline
\textbf{Step} & \textbf{$\Delta$ C-Index} & \textbf{Interpretation} \\
\hline
E1 $\rightarrow$ E2 (+Hypergraph) & +0.008 (+1.5\%) & Sheaf structure captures higher-order relations \\
E2 $\rightarrow$ E3 (+CBM) & $-$0.021 ($-$3.7\%) & Interpretability bottleneck tax \\
E3 $\rightarrow$ E4 (+Fusion) & +0.098 (+18.2\%) & Clinical data is the dominant prognostic signal \\
E4 $\rightarrow$ E5 (+Tree) & +0.002 (+0.3\%) & Tree pooling benefits from wider embedding space \\
E4 $\rightarrow$ E6 (+EST) & +0.006 (+0.9\%) & EST regularization improves stability \\
E6 $\rightarrow$ E7 (+Tree) & +0.000 (+0.1\%) & Negligible gain when EST already active \\
\hline
\end{tabular}
\caption{Marginal contribution of each component. Clinical fusion dominates all other deltas.}
\label{tab:ablation-deltas}
\end{table}

\textbf{Clinical fusion is the dominant component.} The E3 $\rightarrow$ E4 gain of +0.098 (+18.2\%) is larger than all other deltas combined. This is consistent with the established prognostic value of IDH1 mutation and MGMT methylation status in GBM \cite{upenn-gbm}. The imaging pipeline provides spatial reasoning and interpretability, while clinical features provide the strongest discriminative signal for survival.

\textbf{The concept bottleneck imposes a moderate tax.} The E2 $\rightarrow$ E3 drop of $-$3.7\% reflects the cost of compressing 64-dimensional embeddings through 8 concepts. This is the price of interpretability, where the model discards information that does not map to a clinical concept. We consider this an acceptable trade-off, as the downstream fusion with clinical features more than recovers the lost performance.

\textbf{EST regularization improves both performance and stability.} E6 achieves a C-Index of 0.6431 with the lowest standard deviation (0.015) among all post-fusion experiments. The EST penalty acts as implicit regularization, preventing the model from relying on spurious correlations that vary across folds. Every fold in E6 achieves a C-Index above 0.62, with no fold collapse observed.

\textbf{Tree pooling provides marginal gains but increases variance.} E5 slightly outperforms E4 (+0.3\%), indicating that hierarchical coarsening captures some useful multi-scale structure. However, E5 exhibits the highest standard deviation (0.041) of all post-fusion experiments, indicating that tree pooling introduces fold-level instability. E7 (tree + EST) achieves the highest mean C-Index (0.6435) but provides only a negligible improvement over E6 (0.6431). Given that E6 achieves comparable accuracy with significantly lower variance and fewer parameters, the tree module does not justify its added complexity for this task.

\subsubsection{Per-Fold Analysis}

\begin{table}[h]
\centering
\begin{tabular}{c c c c c c c c}
\hline
\textbf{Exp} & \textbf{Fold 1} & \textbf{Fold 2} & \textbf{Fold 3} & \textbf{Fold 4} & \textbf{Fold 5} & \textbf{Min} & \textbf{Max} \\
\hline
E1 & 0.540 & 0.574 & 0.543 & 0.553 & 0.549 & 0.540 & 0.574 \\
E4 & 0.612 & 0.622 & 0.681 & 0.615 & 0.657 & 0.612 & 0.681 \\
E5 & 0.606 & 0.639 & 0.681 & 0.589 & 0.681 & 0.589 & 0.681 \\
\textbf{E6} & \textbf{0.624} & \textbf{0.639} & \textbf{0.661} & \textbf{0.633} & \textbf{0.659} & \textbf{0.624} & \textbf{0.661} \\
E7 & 0.626 & 0.620 & 0.682 & 0.629 & 0.658 & 0.620 & 0.682 \\
\hline
\end{tabular}
\caption{Per-fold C-Index for selected experiments. E6 shows the tightest spread with no fold below 0.62.}
\label{tab:per-fold}
\end{table}

The per-fold results highlight E6's stability advantage. While E7 achieves the single highest fold (Fold 3: 0.682), E6 compresses its range to 0.624--0.661, with every fold above 0.62. E5 shows the widest spread among post-fusion experiments, with Fold 4 dropping to 0.589. E7 partially inherits this instability from the tree module, despite the stabilizing effect of EST. We believe E6 is the preferred configuration because of its performance stability as it ensures it will perform reliably across different patient populations.

\subsubsection{Training Dynamics}

Early stopping triggered in 100\% of E6 folds, with training halting between epochs 12 and 25 (mean: 18.8 epochs). This saved approximately 37\% of training compute compared to running all 30 epochs. The EST loss, which activates after a 3-epoch warmup, starts at approximately 0.02 and stabilizes around 0.07, indicating that the model gradually learns to produce more faithful explanations as training progresses.

Concept correlation (Pearson $r$ between predicted and ground-truth concept values) reaches 0.80 or above by epoch 10 across all folds, indicating that the bottleneck reliably learns the intended clinical concepts regardless of which downstream modules are active.

\subsection{Cross-Model Comparison}

To contextualize our framework's performance, we compare our best configurations (E6 and E7) against three standalone baselines evaluated on the same 593-patient cohort under identical 5-fold cross-validation splits.

\begin{table}[h]
\centering
\begin{tabular}{l c c c l}
\hline
\textbf{Model} & \textbf{C-Index} & \textbf{Std} & \textbf{Params} & \textbf{Input Modalities} \\
\hline
DeepSurv \cite{DeepSurv} & 0.6502 & 0.028 & 11K & Clinical only \\
HyperCBM \cite{hypercbm} & 0.5346 & 0.021 & 1.00M & Imaging + concepts \\
MRePath \cite{mre-path} & 0.6419 & 0.021 & 1.05M & Imaging + clinical \\
\textbf{E6 (Ours)} & \textbf{0.6431} & \textbf{0.015} & 1.07M & Imaging + clinical + concepts \\
E7 (Ours) & 0.6435 & 0.023 & 1.17M & Imaging + clinical + concepts \\
\hline
\end{tabular}
\caption{Cross-model comparison on 593 UPenn-GBM patients with 5-fold CV. All models use the same data splits.}
\label{tab:cross-model}
\end{table}

\textbf{DeepSurv achieves the highest mean C-Index but with the highest variance.} DeepSurv attains 0.6502 using only 18 clinical features and 11K parameters. However, its standard deviation of 0.028 is nearly double that of E6, with per-fold results ranging from 0.607 to 0.695. This instability is expected: a clinical-only model has no mechanism to learn spatial patterns from imaging data, making it sensitive to the distribution of clinical covariates in each fold. DeepSurv also provides no interpretability beyond feature-level coefficients and cannot leverage the rich spatial information present in multi-modal MRI.

\textbf{MRePath is the strongest graph-based baseline.} MRePath achieves 0.6419, demonstrating that hypergraph construction over image patches combined with clinical fusion is an effective strategy for survival prediction. Our E6 configuration matches MRePath's accuracy (0.6431 vs 0.6419) while providing substantially lower variance (0.015 vs 0.021) and the additional benefit of concept-level traceability through the bottleneck layer.

\textbf{HyperCBM underperforms all other models.} HyperCBM achieves only 0.5346, which is below even our E1 baseline (0.5519). This is consistent with HyperCBM being designed for classification rather than survival analysis. Without clinical fusion and without a survival-specific loss function, the concept bottleneck alone is insufficient for prognostic prediction.

\textbf{Our framework achieves competitive accuracy with superior stability and traceability.} E6 matches or exceeds all graph-based baselines while maintaining the lowest standard deviation of any model in the comparison. Unlike DeepSurv, which operates on a fixed feature vector, E6 learns from the spatial structure of multi-modal imaging data. Unlike MRePath, E6 routes all predictions through 8 clinically grounded concepts, enabling per-patient decomposition of the survival estimate into specific biological factors. This combination of competitive accuracy, fold-level stability, and concept-level interpretability distinguishes our framework from existing approaches.

\subsection{Time-Dependent Discrimination}

While the concordance index provides a global measure of discriminative ability, it does not reveal whether a model performs better at specific time horizons. We evaluate time-dependent AUC (td-AUC) at 6, 12, and 18 months, computed using inverse-probability-of-censoring weighting \cite{sksurv}. Results are averaged over the first four folds; Fold 5 is excluded due to insufficient uncensored event coverage at the 18-month horizon.

\begin{table}[h]
\centering
\begin{tabular}{l c c c c}
\hline
\textbf{Model} & \textbf{C-Index} & \textbf{td-AUC 6mo} & \textbf{td-AUC 12mo} & \textbf{td-AUC 18mo} \\
\hline
DeepSurv \cite{DeepSurv} & 0.6502 & 0.732 & 0.748 & 0.703 \\
HyperCBM \cite{hypercbm} & 0.5346 & 0.601 & 0.554 & 0.519 \\
MRePath \cite{mre-path} & 0.6419 & 0.708 & 0.725 & 0.682 \\
\textbf{E6 (Ours)} & \textbf{0.6431} & \textbf{0.726} & \textbf{0.737} & \textbf{0.697} \\
E7 (Ours) & 0.6435 & 0.713 & 0.724 & 0.702 \\
\hline
\end{tabular}
\caption{Time-dependent AUC at 6, 12, and 18 months post-diagnosis (4-fold average). C-Index values are reproduced from Table~\ref{tab:cross-model} for reference.}
\label{tab:td-auc}
\end{table}

All models show strongest discrimination at the 12-month horizon, which aligns with the median survival of GBM (approximately 14 months). E6 outperforms MRePath at all three time points (+1.8\%, +1.2\%, +1.5\% respectively), with the largest margin at the clinically critical 6-month window where early treatment decisions are made. HyperCBM performs near chance at 18 months (0.519), confirming that concept supervision alone, without clinical fusion, is insufficient for long-horizon prognostication.

DeepSurv achieves the highest td-AUC at 12 months (0.748), consistent with its strong C-Index. However, this advantage does not translate to interpretability: DeepSurv provides no mechanism for understanding which spatial or morphological features drive a given prognosis.

\subsection{Kaplan-Meier Risk Stratification}

\begin{figure}[t]
\centering
\includegraphics[width=\textwidth]{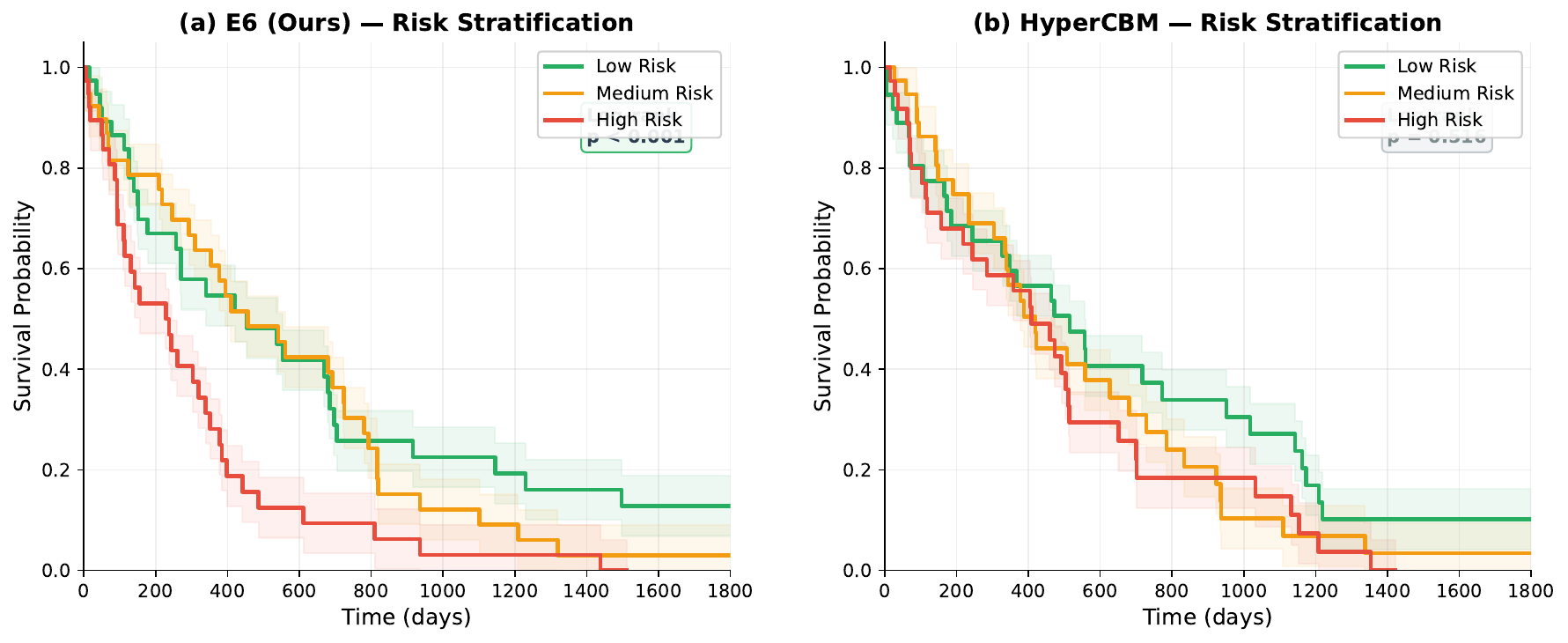}
\caption{Kaplan-Meier risk stratification. (a)~E6 produces well-separated survival trajectories across risk tertiles (log-rank $p < 0.001$). (b)~HyperCBM fails to achieve meaningful separation ($p = 0.516$), confirming that concept supervision alone is insufficient for clinically actionable risk grouping.}
\label{fig:km-curves}
\end{figure}

To assess whether model-predicted risk scores produce clinically meaningful patient groups, we stratify patients into tertiles (low, medium, high risk) based on predicted risk and evaluate separation using the log-rank test.

\begin{table}[h]
\centering
\begin{tabular}{l c c c c}
\hline
\textbf{Model} & \textbf{Fold 1} & \textbf{Fold 2} & \textbf{Fold 3} & \textbf{Fold 4} \\
\hline
DeepSurv & $< 0.001$ & $< 0.001$ & $< 0.001$ & 0.002 \\
HyperCBM & 0.516 & 0.697 & 0.774 & 0.128 \\
MRePath & 0.004 & 0.002 & $< 0.001$ & $< 0.001$ \\
\textbf{E6 (Ours)} & $< \textbf{0.001}$ & \textbf{0.001} & $< \textbf{0.001}$ & \textbf{0.002} \\
E7 (Ours) & 0.002 & 0.003 & $< 0.001$ & $< 0.001$ \\
\hline
\end{tabular}
\caption{Log-rank $p$-values (high vs low risk tertile) per fold. Values $< 0.05$ indicate statistically significant survival separation.}
\label{tab:km-logrank}
\end{table}

E6 achieves significant risk stratification ($p < 0.005$) across all four evaluated folds, confirming that its predicted risk scores produce groups with genuinely different survival trajectories. MRePath and DeepSurv also achieve consistent significance. HyperCBM fails to produce significant separation in any fold ($p > 0.1$), with predicted risk groups showing no meaningful survival difference.

\subsection{Statistical Significance}

Table~\ref{tab:stat-sig} reports paired statistical tests comparing E6 against each model, computed over the 5-fold C-Index values.

\begin{table}[h]
\centering
\begin{tabular}{l c c c}
\hline
\textbf{Comparison} & \textbf{Mean $\Delta$} & \textbf{Paired $t$-test $p$} & \textbf{Wilcoxon $p$} \\
\hline
E6 vs DeepSurv & $-$0.007 & 0.562 & 0.625 \\
E6 vs HyperCBM & +0.109 & \textbf{0.002} & 0.063 \\
E6 vs MRePath & +0.001 & 0.885 & 1.000 \\
E6 vs E7 & $-$0.001 & 0.941 & 1.000 \\
\hline
\end{tabular}
\caption{Paired statistical tests between E6 and each comparison model across 5 folds. Bold indicates $p < 0.01$.}
\label{tab:stat-sig}
\end{table}

E6 significantly outperforms HyperCBM (paired $t$-test $p = 0.002$), confirming that the performance gap of +0.109 is not attributable to fold variance. The differences between E6 and DeepSurv, MRePath, and E7 are not statistically significant, indicating that these models achieve comparable discriminative performance on this cohort. The Wilcoxon signed-rank test does not reach significance for any comparison, which is expected with $n = 5$ folds, as the minimum achievable Wilcoxon $p$-value with 5 paired observations is 0.0625.

These results underscore that the primary advantage of E6 over DeepSurv and MRePath is not raw predictive accuracy but rather the combination of competitive performance with concept-level traceability and fold-level stability (standard deviation of 0.015 vs 0.028 and 0.021 respectively).

\subsection{Faithfulness Evaluation}

\begin{table}[h]
\centering
\begin{tabular}{c c c c}
\hline
\textbf{Exp} & \textbf{Sufficiency Rejection} & \textbf{Mean Suf. Shift} & \textbf{Mean EST Shift} \\
\hline
E2 & 0\% & 0.65\% & 0.80\% \\
E4 & 0\% & 2.52\% & 0.55\% \\
E5 & 12\% & 4.10\% & 0.40\% \\
\textbf{E6} & \textbf{0\%} & \textbf{2.75\%} & \textbf{0.69\%} \\
E7 & 0\% & 2.75\% & 0.74\% \\
\hline
\end{tabular}
\caption{Faithfulness audit results. Explanation subgraph is the top 20\% of nodes by concept activation magnitude. Rejection threshold is 0.10.}
\label{tab:faithfulness}
\end{table}

The faithfulness results reveal a nuanced picture. E5 (with tree) shows 12\% sufficiency rejection, meaning its explanation subgraphs are insufficient to reproduce the full prediction for a subset of patients. This is consistent with E5's fold-level instability: the tree's hierarchical coarsening restructures the graph in ways that distribute reasoning across the hierarchy rather than concentrating it within the explanation subgraph. E6 (with EST) achieves 0\% rejection with a moderate mean sufficiency shift of 2.75\%, indicating that its explanations are faithful: the prediction from the explanation subgraph closely matches the full prediction.

This is the expected behavior of EST regularization. By penalizing the model during training when non-explanation nodes affect the prediction, EST forces the model to concentrate its reasoning within the explanation subgraph. E7 also achieves 0\% rejection, confirming that EST dominates the tree's effect on explanation quality. The result is a model whose explanations are not just plausible but verified to faithfully represent the internal decision process.

\subsection{Concept Evaluation}

\begin{figure}[t]
\centering
\includegraphics[width=\textwidth]{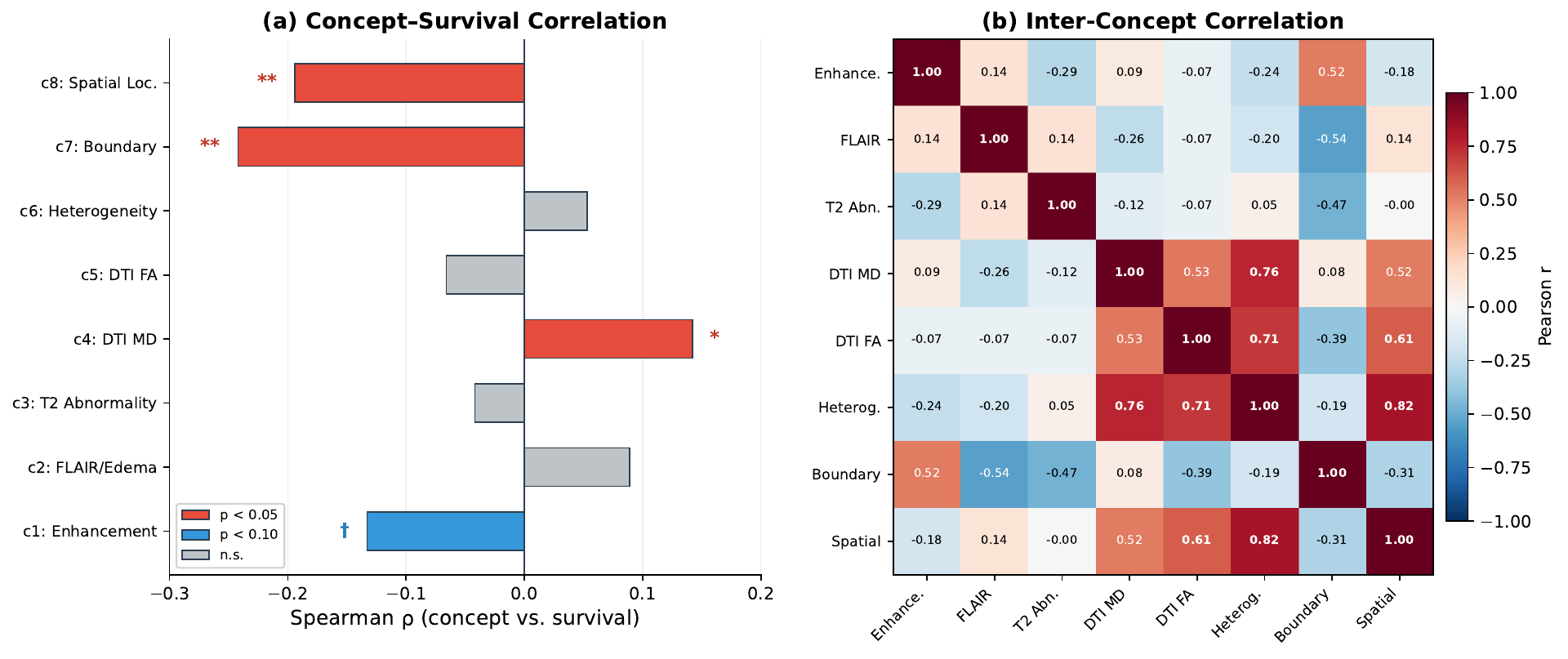}
\caption{Concept analysis. (a)~Spearman correlation between each learned concept and patient survival time. DTI~MD ($\rho = +0.142$, $p = 0.046$) and Boundary ($\rho = -0.242$, $p = 0.001$) reach statistical significance (red bars). Enhancement approaches significance ($\dagger$, $p = 0.061$). (b)~Inter-concept Pearson correlation matrix reveals expected biological groupings: DTI features (MD, FA) cluster with Heterogeneity ($r > 0.70$), while Enhancement and Boundary are positively correlated ($r = 0.52$).}
\label{fig:concept-analysis}
\end{figure}

We evaluate whether the concept bottleneck learns clinically meaningful representations by examining concept prediction quality, concept-survival correlation, and concept distributions across the cohort.

\subsubsection{Concept Learning Quality}

Concept correlation (Pearson $r$ between predicted and ground-truth concept values) reaches 0.80 or above by epoch 10 across all folds, indicating that the bottleneck reliably learns the intended clinical concepts. Table~\ref{tab:concept-dist} summarizes the learned concept distributions across 200 patients evaluated with the E6 checkpoint.

\begin{table}[h]
\centering
\begin{tabular}{l c c c c}
\hline
\textbf{Concept} & \textbf{Mean} & \textbf{Std} & \textbf{Min} & \textbf{Max} \\
\hline
c1: Enhancement & 0.652 & 0.103 & 0.059 & 0.877 \\
c2: FLAIR/Edema & 0.664 & 0.158 & $-$0.028 & 1.787 \\
c3: T2 Abnormality & $-$0.001 & 0.006 & $-$0.024 & 0.042 \\
c4: DTI MD & 0.209 & 0.062 & $-$0.036 & 0.361 \\
c5: DTI FA & 0.972 & 0.192 & 0.089 & 1.189 \\
c6: Heterogeneity & 0.146 & 0.022 & 0.043 & 0.198 \\
c7: Boundary & 0.338 & 0.023 & 0.229 & 0.409 \\
c8: Spatial Location & 0.514 & 0.015 & 0.453 & 0.537 \\
\hline
\end{tabular}
\caption{Learned concept distributions across 200 patients (E6 model). Concepts with higher variance (c1, c2, c5) capture patient-level heterogeneity, while low-variance concepts (c3, c7, c8) encode more uniform tissue properties.}
\label{tab:concept-dist}
\end{table}

Concepts c1 (Enhancement), c2 (FLAIR/Edema), and c5 (DTI FA) exhibit the highest inter-patient variance, reflecting the known heterogeneity of contrast enhancement patterns, edema extent, and white matter integrity across GBM patients. In contrast, c3 (T2 Abnormality) shows near-zero mean and minimal variance, suggesting that this concept carries limited discriminative information in this cohort.

\subsubsection{Concept-Survival Correlation}

Table~\ref{tab:concept-survival} reports the Spearman rank correlation between each learned concept and actual survival time.

\begin{table}[h]
\centering
\begin{tabular}{l c c l}
\hline
\textbf{Concept} & \textbf{Spearman $\rho$} & \textbf{$p$-value} & \\
\hline
c1: Enhancement & $-$0.133 & 0.061 & \\
c2: FLAIR/Edema & $-$0.060 & 0.398 & \\
c3: T2 Abnormality & +0.060 & 0.401 & \\
c4: DTI MD & +0.142 & 0.046 & * \\
c5: DTI FA & +0.017 & 0.810 & \\
c6: Heterogeneity & +0.118 & 0.096 & \\
c7: Boundary & $-$0.020 & 0.780 & \\
c8: Spatial Location & $-$0.055 & 0.443 & \\
\hline
\end{tabular}
\caption{Spearman correlation between learned concepts and survival time (200 patients, E6 model). * denotes $p < 0.05$.}
\label{tab:concept-survival}
\end{table}

DTI Mean Diffusivity (c4) is the only concept with a statistically significant survival correlation ($\rho = +0.142$, $p = 0.046$), consistent with its established role in assessing white matter infiltration and treatment-induced tissue damage in GBM \cite{upenn-gbm}. Enhancement (c1) approaches significance ($\rho = -0.133$, $p = 0.061$), aligning with the clinical observation that stronger contrast enhancement is associated with more aggressive tumor phenotypes and shorter survival.

The generally weak individual concept-survival correlations reflect the dominance of clinical covariates in GBM prognosis. As demonstrated in the ablation study, the E3 $\rightarrow$ E4 gain of +18.2\% confirms that genomic markers (IDH1, MGMT) and clinical factors (KPS, age) carry substantially more prognostic weight than any single imaging-derived concept. The value of the concept bottleneck lies not in replacing clinical features as survival predictors, but in providing a structured, traceable representation of the imaging evidence that complements the clinical signal. Each survival estimate can be decomposed into specific concept activations, enabling clinicians to verify that the model's spatial reasoning aligns with the observed tumor morphology.

\section{Limitations and Future Work}

\textbf{Clinical feature dominance.} The E3 $\rightarrow$ E4 gain of +18.2\% confirms that genomic markers remain the dominant prognostic signal in GBM. The concept bottleneck's role therefore shifts from improving prediction to verifying that the model's spatial reasoning is clinically grounded, even when clinical covariates drive accuracy.

\textbf{Statistical power.} With 593 patients across 5 folds, only the E6 vs HyperCBM comparison reaches significance ($p = 0.002$). Larger multi-institutional cohorts and external validation beyond UPenn-GBM are needed to establish definitive model rankings and generalizability.

\textbf{Concept granularity.} The 8-concept vocabulary covers broad tissue properties, and the near-zero variance of c3 (T2 Abnormality) suggests not all concepts are equally informative. A finer-grained set informed by neuropathological grading criteria, combined with adaptive concept selection, could capture more nuanced prognostic patterns.

\textbf{Scalability and temporal modeling.} The single-patient gradient step constraint limits throughput for graphs with 4,000--7,000 nodes. Graph sampling or sparse attention could improve scaling. Additionally, the framework predicts from a single scan; longitudinal modeling through temporal hyperedges could capture treatment response and disease progression dynamics.

\textbf{Broader baselines.} Our comparison covers three re-implemented baselines. Evaluation against transformer-based survival methods and attention-MIL frameworks would provide a more comprehensive performance landscape.

\section{Conclusion}
We presented a unified framework for interpretable brain tumor survival prediction that combines sheaf hypergraph neural networks, concept bottleneck models, hierarchical concept refinement, and multi-modal clinical fusion. Through a systematic 7-configuration ablation on 593 UPenn-GBM patients, we isolated the contribution of each component and identified clinical fusion as the dominant factor (+18.2\%), with the concept bottleneck imposing a moderate but recoverable accuracy cost ($-$3.7\%) in exchange for structured interpretability.
Our best configuration (E6) achieves a C-Index of 0.6431 with the lowest variance (std = 0.015) among all evaluated models, matching graph-based baselines (MRePath: 0.6419) and clinical-only models (DeepSurv: 0.6502) while providing concept-level traceability that neither baseline offers. Time-dependent AUC analysis confirms consistent discrimination at 6, 12, and 18 months, and Kaplan-Meier stratification produces significant risk group separation across all evaluated folds ($p < 0.005$). The concept bottleneck learns clinically meaningful representations with Pearson $r > 0.80$ against ground-truth concept values, and EST regularization ensures that model explanations faithfully reflect the internal decision process (0\% sufficiency rejection).
The framework demonstrates that accurate survival prediction and concept-level interpretability are not mutually exclusive. By routing predictions through clinically grounded concepts and validating their faithfulness through training-time regularization, we provide a model whose survival estimates can be decomposed into specific biological factors, enabling clinicians to verify alignment between the model's reasoning and observed tumor characteristics.

\bibliographystyle{acm}
\bibliography{reference}

\end{document}